\documentclass[letterpaper, conference]{IEEEtran}
\IEEEoverridecommandlockouts
\usepackage{times}
\usepackage{epsfig}
\usepackage{graphicx}
\usepackage{amsmath}
\usepackage{amssymb}
\usepackage{tabularx} 
\usepackage{enumitem} 
\usepackage{dirtytalk}
\usepackage{comment}
\usepackage{csquotes}

\usepackage{caption}
\usepackage{todonotes}
\begin{document}

\title{Towards Sensor Data Abstraction of Autonomous Vehicle Perception Systems\\
\thanks{This work results from the KI Data Tooling project supported by the Federal Ministry for Economic Affairs and Energy (BMWi), grant numbers 19A20001I, 19A20001J, 19A20001L and 19A20001O.}
}

\author{\IEEEauthorblockN{Hannes Reichert\IEEEauthorrefmark{2}\textsuperscript{\textasteriskcentered},
Lukas Lang\IEEEauthorrefmark{3}\textsuperscript{\textasteriskcentered},
Kevin Rösch\IEEEauthorrefmark{4}\textsuperscript{\textasteriskcentered},
Daniel Bogdoll\IEEEauthorrefmark{4}\textsuperscript{\textasteriskcentered},\\ 
Konrad Doll\IEEEauthorrefmark{2}, 
Bernhard Sick\IEEEauthorrefmark{5}, 
Hans-Christian Reuss\IEEEauthorrefmark{3}, 
Christoph Stiller\IEEEauthorrefmark{4}
and J. Marius Zöllner\IEEEauthorrefmark{4}}

\IEEEauthorblockA{\IEEEauthorrefmark{2}University of Applied Sciences Aschaffenburg, Germany\\
Email: \{hannes.reichert, konrad.doll\}@th-ab}
\IEEEauthorblockA{\IEEEauthorrefmark{3}IFS at University of Stuttgart, Germany\\
Email: \{lukas.lang, hans-christian.reuss\}@ifs.uni-stuttgart.de}
\IEEEauthorblockA{\IEEEauthorrefmark{4}FZI Research Center for Information Technology, Germany\\
Email: \{kevin.roesch, bogdoll, stiller, zoellner\}@fzi.de}
\IEEEauthorblockA{\IEEEauthorrefmark{5}University of Kassel, Germany\\
Email: bsick@uni-kassel.de}}

\maketitle
\begingroup\renewcommand\thefootnote{\textasteriskcentered}
\footnotetext{These authors contributed equally}
\endgroup

\begin{abstract}
Full-stack autonomous driving perception modules usually consist of data-driven models based on multiple sensor modalities. However, these models might be biased to the sensor setup used for data acquisition. This bias can seriously impair the perception models' transferability to new sensor setups, which continuously occur due to the market's competitive nature. We envision sensor data abstraction as an interface between sensor data and machine learning applications for highly automated vehicles (HAD).

For this purpose, we review the primary sensor modalities, camera, lidar, and radar, published in autonomous-driving related datasets, examine \emph{single sensor abstraction} and \emph{abstraction of sensor setups}, and identify critical paths towards an abstraction of sensor data from multiple perception configurations.
\end{abstract}


\section{Introduction}
Designing and mounting a sensor setup for a research vehicle requires much time and engineering effort. 
Furthermore, the amount of data needed for highly automated driving is soaring, and the acquisition and annotation of sensor data are expensive and time-consuming, yielding a limited collection of publicly available data sets used for HAD research. In addition, most data sets consist only of data acquired with a fixed sensor setup.
\begin{figure}[!t] 
    \centering
    \includegraphics[width=\columnwidth]{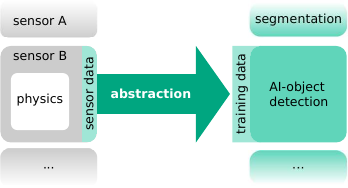}
    \caption{Sensor data abstraction provides an interface between a sensor and further applications.}
    \label{fig:sens_dat_abs}
\end{figure}

The properties of the sensor or sensor setup can implicitly bias the data and methods developed on it. This bias might reduce the transferability of methods and algorithms to new sensor setups. This bias might reduce the transferability of methods and algorithms to new sensor setups.
In this work, we want to explore the benefit of using sensor data abstraction in HAD. As displayed in fig. \ref{fig:sens_dat_abs}, we see sensor data abstraction as an interface between sensor data and machine learning applications, improving existing models' performance and utility. We envision a unified and sensor-independent abstraction of sensor data, enabling a broad perception pipeline for different sensor setups.

In the following chapter \ref{sec: SOTA}, we will introduce state-of-the-art sensor setups in HAD-related datasets. In section \ref{sec:towards_sensor_abstraction} we examine the abstraction of data from single sensors, which we refer to as \emph{marginal sensor data abstraction} and the abstraction of data from sensor setups, which we refer to as \emph{joint sensor data abstraction}. Finally, section \ref{sec:outlook} summarizes our findings and points towards challenges and further research directions.

\section{State of the Art} \label{sec: SOTA}

The following section will first introduce several open-access datasets. They provide a basis for the methods presented in this paper. After that, different sensor modalities and their characteristics will be discussed. At the end of this chapter, sec. \ref{sec:degrees_of_abstraction} will introduce the field of sensor data abstraction and standard definitions.

\subsection{Sensor Setups in Datasets} \label{sec:datasets}

For the abstraction of sensor data, it is necessary to understand the implicit properties of the utilized sensors and their setup. These have a strong influence on the type and quality of the resulting data. For this purpose, we examine popular HAD-related datasets as listed in \cite{fen2021DeepMultimodal}. 
While camera images and lidar point clouds are typically included, only recently datasets included radar data
Sensor data is usually decoupled from the physical sensor by a hardware abstraction layer (raw sensor data) and provided as point clouds or images. This hardware-related signal processing, whose resulting representation contains implicit sensors' characteristics, is not examined in this paper. In the following, we will only refer to this resulting representation as \emph{sensor data}, shown in fig. \ref{fig:sens_dat_abs}.

\subsection{Sensor Modalities} \label{sec:sensor_modalities}

\begin{figure}
\includegraphics{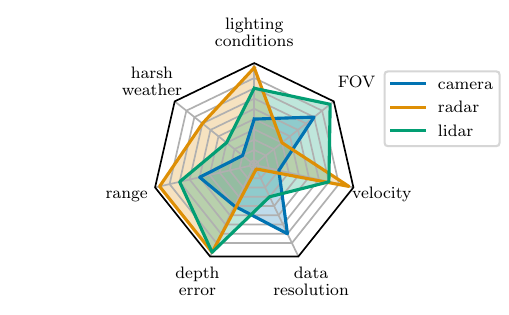}
\centering
\caption{Quantitative comparison between the three major sensor modalities. From bad (center) to excellent (rim).}
\label{fig: compare_sensors}
\end{figure}

There are common characteristics, such as position and orientation, that differ for each data collection process and affect all sensor types. The influence of other common features, such as resolution, the field of view (FOV), and manufacturing tolerances, depending on the sensor configuration. A sensor's characteristics are implied in its data representation, biasing the method used to process the data. This risk exists with both learned and engineered methods. An example of this bias can be found in \cite{DBLP:journals/corr/DoerschGE15}. The authors tried to find the relative position of corresponding image patches with a convolutional neural network (CNN). However, the CNN seems to learn a trivial solution by locating the patches based on the chromatic aberration implied to the image due to lens characteristics. The sensor characteristics lead to a specific profile which is shown in fig. \ref{fig: compare_sensors} for camera, lidar and radar.

Based on \cite{rosique2019systematic} we define seven parameters to visualize the strengths and weaknesses inside that profile. 
The values are based on ideal sensors. To determine the possible marginal, abstract spaces, one can use the filled areas inside the plot. 
As seen in the figure, the sum of all characteristics also results in a shared space. This space could be covered in a sensor fusion and, therefore, in a shared (joint) abstraction. For further research, both spaces for marginal and joint abstraction need to be considered.

\subsection{Sensor Data Abstraction} \label{sec:degrees_of_abstraction}
Kirman et al. first coined the term \emph{abstract sensor} as a module that provides a \say{mapping from the actual sensor space to the observation space} \cite{kir1991SensorAbstractionsa}. As their observation space is well designed, the resulting \say{abstract sensor reading} is human-interpretable. Jha et al. provide a similar understanding of a \say{sensor abstraction layer}, which performs several preprocessing steps of input data before given to the perception layer \cite{jha2019MLBasedFault}.

From a software development perspective, sensor abstraction is often understood as a consistent and unified interface for many sensors, which can have different sensing technologies or are made from different companies \cite{kab2006VirtualSensors} 
. This concept neglects the implicit information included in the sensor setup but might provide it as metadata.

In contrast to classical, hand-engineered abstractions, such as mathematical transformations, in Machine Learning, neural networks are frequently used to learn latent spaces, abstractions of input data. These are typically not directly human-interpretable. Latent spaces can be utilized for specific perception tasks, such as object detection or state representation for end-to-end systems \cite{che2020InterpretableEndtoend}.

\section{Towards Sensor Abstraction}
\label{sec:towards_sensor_abstraction}
As shown in the previous sec. \ref{sec:degrees_of_abstraction}, there exist a multitude of meanings for sensor (data) abstraction in the literature. We aim to provide insights into sensor-independent and unified environment representations for modern perception approaches. Thus we define the term as:

\begin{displayquote}
\emph{Sensor data abstraction is a learned mapping of sensor data from $1-n$ sensors to a unified abstraction suitable as input data for neural networks}
\end{displayquote}

Based on this definition, which is also shown in fig. \ref{fig:sens_dat_abs}, we will discuss the abstraction of single sensors, sensor setups and informed abstraction based on meta-data in the following chapter. The latter are highly motivated by the sensor's characteristics shown in fig. \ref{fig: compare_sensors}.

\subsection{Marginal Sensor Data Abstraction} 
\label{sec: marginal}

As mentioned in sec. \ref{sec:sensor_modalities} the implicit encoding of sensor parameters to the representation is a problem. For camera data, as an example, this problem can be partially solved by empiric models, optimized on large-scale datasets with a wide variety of used cameras.
In detail, such CNN-based models often consist of a two-stage architecture \cite{Jiao_2019}. Such two-stage architectures include a \emph{backbone stage}, which encodes the sensor data to an abstract latent space, and a \emph{head stage}, which uses the extracted features to perform tasks like classification, object detection, or human pose estimation segmentation.
From vast datasets like COCO \cite{lin2015microsoft}, backbones can learn to extract generalized features such that transferring to new data or tasks can be done by retraining.
A model trained on COCO can avoid sensor bias due to the sheer amount of images taken from various perspectives with multiple consumer grade cameras.

Unlike cameras, lidar and radar sensors represent the environment as point clouds, i. e. a radar generates data with additional values like velocity and magnitude. In literature there are various methods, e.g. PointNet \cite{PointNet}, capable of directly working on 3D data.

In summary, an abstraction of various sensors can be achieved empirically inter alia with state-of-the-art two-stage perception models, given enough and variable data. For HAD datasets, this is currently not the case. Therefore, one core question is how to acquire suitable, sensor-specific datasets regarding scale and variety, given that marginal sensor abstraction architectures are relatively mature at this point.

\subsection{Joint Sensor Data Abstraction}\label{sec: joint}

Besides marginal sensor abstractions as discussed in sec. \ref{sec: marginal}, we want to outline the necessity and feasibility of a joint sensor abstraction and highlight early results in this research field. As the authors in \cite{fay2020DeepLearning} demonstrate, multi-modal sensor setups are needed to provide a safe perception pipeline for HAD. As an example, camera-only sensor setups are vulnerable to harsh weather, which can be compensated with additional sensor(s), as seen in fig. \ref{fig: compare_sensors}.

We can see joint abstraction as a subset of \emph{sensor fusion}. While most fusion approaches are based on a high level of data abstraction, joint sensor abstraction focuses on \emph{low-level} fusion where raw sensor data from multiple sensors is fused. While this approach contains the most information, it is also extremely complex \cite{ott2013FusionData}
. Nobis et al. have also shown that it is possible to learn the optimal level for fusion \cite{nobis2020deep}.

The datasets mentioned in sec. \ref{sec:datasets} act as a basis for these approaches. They were recorded by research vehicles with certain sensor setups regarding the number, types, manufacturers, positions, and orientations of the utilized sensors. Such configurations can be split into three categories: \emph{Complementary} for the combination of independent sensors, \emph{competitive} for reliability and \emph{cooperative} for enhanced quality by integrating higher-level measurements which cannot be measured directly \cite{dur1988SensorModels}. Information about such configurations is implicitly included in the sensor data and also explicitly available as meta-data, which we will examine in sec. \ref{sec:metadata}. An example of a \emph{complementary} is the usage of multiple cameras, each observing disjunctive parts of the world. In \emph{competitive} setups, the output of multiple, redundant sensors can be used as input for a voting model. Examples for \emph{cooperative} would be a mapping of camera data to point clouds or a transformation of point clouds onto an image representation (e. g., RGB-Depth) before a marginal abstraction. This mapping of one sensor into the representation of another gives us a \textit{novel sensor}. This \textit{novel sensor} contains information sensed by its parental sensors but novel characteristics. If the marginal abstraction is done as described in sec. \ref{sec: marginal}, an empiric model has the freedom to decide which components of this novel sensor representation are helpful.

Based on \emph{mid-level} and \emph{high-level} fusion approaches and \emph{competitive} as well as \emph{cooperative} configurations, many abstraction architectures can be engineered. Examples are mapping of multi-modal sensor data onto a common representation, e. g., radar and lidar onto image formats, merging posterior to a marginal abstraction, or pure transformations 
\cite{nobis2020deep}. Due to their nature of equalized input data representation, such approaches are currently prevalent for learned abstraction approaches, as Cui et al. demonstrate \cite{cui2021DeepLearning}. Famous examples are end-to-end models, which encode multi-modal sensor data, mostly camera and lidar, into an abstract latent space to perform the complete driving task based on this environment representation. Che et al. present a method that shows the quality of the latent space by constructing a birds-eye view based on it \cite{che2020InterpretableEndtoend}. Jointly trained variational autoencoders follow similar research directions \cite{kor2019JointlyTraineda}.

Due to the high complexity of \emph{low-level} approaches, potentially suitable neural network architectures are not well understood yet based on the different types of input data representation. Also, current approaches suffer under model inflexibility - new sensing modalities typically need full retraining of the network \cite{fen2021DeepMultimodal}. Therefore, further research is necessary.

Regardless of the joint abstraction, the sensors' bias remains, and its impact on the data could be even higher due to coherence's between the sensors in one setup. 

\subsection{Utilizing Metadata for Abstraction} \label{sec:metadata}
So far, we have only dealt with abstraction in terms of sensor data. However, an automotive sensor setup usually comes with auxiliary data describing sensors themselves or geometric relations between multiple sensors. We refer to this data as metadata. Following the concept of \emph{informed machine learning}, this metadata might be helpful for sensor data abstraction \cite{von2019InformedMachine}. An example for this is CamConvs \cite{facil2019camconvs}, which explicitly encodes the intrinsic parameters of cameras to its image representation to overcome a sensor bias, yielding considerable improvements in the task of mono depth estimation when train and test images are acquired with different cameras. The research question arises if and how metadata can be utilized to improve both marginal (intrinsic parameters) and joint (intrinsic and extrinsic parameters) abstraction. There lies much potential in metadata regarding joint abstraction. There is still no consensus among the industry leaders in HAD regarding an optimal sensor setup. Eslami et al. have shown that it is possible to learn a scene representation by providing two different camera perspectives as input and predict the scene from a third camera viewpoint. Based on this integration of metadata, it is possible to utilize the prediction error to analyze optimal positions for additional sensors. A significant error suggests  that the existing sensors could not capture the content from the new viewpoint \cite{esl2018NeuralScene}. Further research for multi-modal approaches is necessary.

\subsection{General pipeline for multiple sensor configurations} \label{sec:general_pipeline}

\begin{figure}[h!]
    \centering
    \includegraphics[width=\columnwidth]{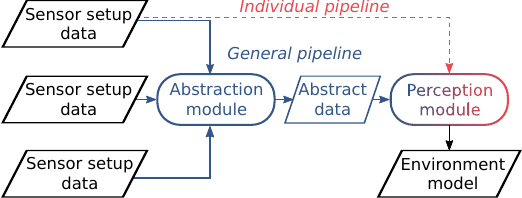}
    \caption{Data flow from sensors to an environment model. 
    The abstraction layer allows for a decoupling of observations and the state of a system. 
    }
    \label{fig:data_flow}
\end{figure}

Based on the current situation of available sensor setups in HAD, the research question arises how to represent multiple sensor setups, such that utilization to new setups is given. As introduced in the beginning, we envision a unified  and  sensor-independent  abstraction  of  sensor  data,  enabling  a  general perception pipeline for different sensor setups as shown in fig. \ref{fig:data_flow}. While there are many open questions, this would enable the transferability of whole perception pipelines between multiple, with different sensor configurations equipped, autonomous vehicles.

Due to the complementary nature of the installed sensors, one cannot expect that it will be possible to utilize arbitrary setups which will lead to the same abstract environment model as all others. Therefore, a core research question is the space, in which the positions as well as the combinations of sensors can be altered without effects on the abstract representation. For this purpose, the shown concepts and research directions from this section will be key contributors to potential solutions regarding this research field.

\section{Conclusion and Outlook}
\label{sec:outlook}

This paper defined the theme of sensor data abstraction for the field of automotive applications. Furthermore, we discovered limitations and multiple critical paths towards sensor data abstraction. Further research could study different ways to realize marginal and joint abstraction concerning utilization and transferability. 
A possible solution to overcome the limitation of available and unbiased data is to build a vast data set with identical trajectories and variations in the sensor setups used. However, this is not feasible. Further research would therefore be to integrate the real sensor setups into simulation environments such as Carla \cite{Dosovitskiy17}. With such a simulation, it would be less effort to generate data than building multiple test vehicles.  The authors will continue their research on this topic. Since such latent spaces have the potential for further applications, we will also focus on topics such as input augmentation, data compression \cite{tol2018GenerativeNeural}, corner case detection \cite{bre2020CornerCases} and prediction \cite{neu2021VariationalAutoencoderBased}.


{\small
\bibliographystyle{ieee}
\bibliography{egbib}
}

\end{document}